\pdfoutput=1

\documentclass[11pt]{article}

\usepackage{naacl2021}

\usepackage{times}
\usepackage{latexsym}
\usepackage{amsmath}

\usepackage[T1]{fontenc}

\usepackage[utf8]{inputenc}

\usepackage{microtype}
\usepackage{dirtytalk}
\usepackage{graphicx}
\title{Industry Scale Semi-Supervised Learning for Natural Language Understanding}

\author{Luoxin Chen\thanks{$^\star$Equal contribution} \\
  Alexa AI \\
  \texttt{luoxchen@amazon.com} \\
  \And
  Francisco Garcia\footnotemark[1] \\
  Alexa AI \\
  \texttt{fgmz@amazon.com} \\
  \And
  Varun Kumar\footnotemark[1] \\
  Alexa AI \\
  \texttt{kuvrun@amazon.com} \\
  \AND
  He Xie\footnotemark[1] \\
  Alexa AI \\
  \texttt{hexie@amazon.com} \\ 
  \And
  Jianhua Lu \\
  Alexa AI \\
  \texttt{jianhual@amazon.com}
  }
\begin{document}
\maketitle
\begin{abstract}
This paper presents a production Semi-Supervised Learning (SSL) pipeline based on the student-teacher framework, which leverages millions of unlabeled examples to improve Natural Language Understanding (NLU) tasks. We investigate two questions related to the use of unlabeled data in production SSL context: 1) how to select samples from a huge unlabeled data pool that are beneficial for SSL training, and 2) how do the selected data affect the performance of different state-of-the-art SSL techniques. We compare four widely used SSL techniques, Pseudo-Label (PL), Knowledge Distillation (KD), Virtual Adversarial Training (VAT) and Cross-View Training (CVT) in conjunction with two data selection methods including committee-based selection and submodular optimization based selection. We further examine the benefits and drawbacks of these techniques when applied to intent classification (IC) and named entity recognition (NER) tasks, and provide guidelines specifying when each of these methods might be beneficial to improve large scale NLU systems.
\end{abstract}


\section{Introduction}
\label{intro}


Voice-assistants with speech and natural language understanding (NLU) are becoming increasingly prevalent in every day life. These systems, such as Google Now, Alexa, or Siri, are able to respond to queries pertaining multiple domains (e.g., music, weather). An NLU system commonly consists of an intent classifier (IC) and named entity recognizer (NER). It takes text input from an automatic speech recognizer and predicts intent and entities. For example, if a user asks ``play lady gaga'', the IC classifies the query to intent of PlayMusic, and the NER classifies ``lady gaga'' as Artist. An important requirement for voice-assistants is the ability to continuously add support for new functionalities, i.e., new intents, or new entity types, while improving recognition accuracy for the existing ones. Having high quality labeled data is the key to achieve this goal. However, obtaining human annotation is an expensive and time-consuming process. 



Semi-Supervised Learning (SSL) provides a framework for utilizing large amount of unlabeled data when obtaining labels is expensive~\cite{Chapelle2006IntroductionTS, blum1998combining, zhou2005tri}. SSL techniques have been shown to improve deep models performance across different machine learning tasks including text classification, machine translation, image classification~\cite{cvt,vat,vat_classification, yalniz2019billion, berthelot2019mixmatch, chen-etal-2020-seqvat}. A common practice to evaluate SSL algorithms is to take an existing labeled dataset and only use a small fraction of training data as labeled data, while treating the rest of the data as unlabeled dataset. Such evaluation, often constrained to the cases when labeled data is scarce, raises questions about the usefulness of different SSL algorithms in a real-world setting~\cite{oliver2018realistic}. 

In voice assistants, we face additional challenges while applying SSL techniques at scale including (1) how much unlabeled data should we use for SSL and how to select unlabeled data from a large pool of unlabeled data? (2)  Most SSL benchmarks make the assumption that unlabeled datasets come from the same distribution as the labeled datasets. This assumption is often violated as, by design, the labeled training datasets also contain synthetic data, crowd-sourced data to represent anticipated usages of a functionality, and unlabeled data often contain a lot of out of domain data. (3) Unlike widely used NLU datasets such as SNIPS~\cite{Coucke2018SnipsVP}, ATIS~\cite{Price1990EvaluationOS}, real-world voice assistant datasets are much larger and have a lot of redundancy because some queries such as \say{turn on lights} might be much more frequent than others.  Due to such evaluation concerns, performance of different SSL techniques in \say{real-world} NLU applications is still in question.

To  address  these  issues, we study three data selection methods to select unlabeled data and evaluate how the selected data affect the performance of different SSL methods on a real-world NLU dataset. This paper provides three contributions: (1) Design of a production SSL pipeline which can be used to intelligently select unlabeled data to train SSL models (2) Experimental comparison of four SSL techniques including, Pseudo-Label, Knowledge Distillation, Cross-View Training, and Virtual Adversarial Training in a real-world voice assistant setting (3) Operational recommendations for NLP practitioners who would like to employ SSL in production setting.

\section{Background}
\label{background}
Semi-Supervised Learning techniques are capable of providing large improvements in model performance with little effort, which could play a crucial role in large scale systems in industry. In supervised learning, given a labeled dataset $\mathcal{D}_l$ composed of input-label pairs $(x,y)$, the goal is to learn a prediction model $f_{\theta}(x)$, with parameters $\theta$, that is able to predict the correct label $y'$ corresponding to a new unseen input instance $x'$. SSL techniques aim to leverage an unlabeled dataset, $\mathcal{D}_u$, to create better performing models than those that could be obtained by only using $\mathcal{D}_l$.

The two widely used SSL methods are: Pseudo-Label (PL), and Knowledge Distillation (KD). In PL, a teacher model trained on labeled data is used to produce pseudo-labels for the unlabeled data set. A student model trained on the union of the labeled and pseudo-labeled data sets, often outperforms the teacher model.~\cite{yarowsky1995unsupervised,mcclosky2006effective}. On the other hand, KD SSL methods do not assign a particular label to an unlabeled instance, but instead consider the whole distribution over the label space~\cite{parthasarathi2019lessons,liu2019improving,aguilar2020knowledge}. In KD, it is hypothesized that leveraging the probability distribution over all labels provides more information than assuming a definitive label belonging to one particular class~\cite{hinton2015distilling}.

In addition to PL and KD, Virtual Adversarial Training (VAT) and Cross-View Training (CVT) have achieved state-of-the-art SSL performance on various tasks including text classification, named entity recognition, and dependency parsing~\cite{vat,cvt,vat_classification,chen-etal-2020-seqvat}. In this paper, we conduct comprehensive experiments and analysis related to these commonly used SSL techniques, and discuss their pros and cons in the industry setting. 

Data selection for SSL has been explored for different tasks including image classification ~\cite{Ding_2018}, NER ~\cite{ji-grishman-2006-data,Ruder_2018}. Model confidence based data selection is a widely used technique for SSL data selection where unlabeled data is selected on the basis of a classifier's confidence. Due to the abundance of unlabeled data in production voice-assistants, model confidence based filtering leads to a very large data pool. To overcome this issue, we study different data selection algorithm which can further reduce the size of unlabeled data. 

 
\section{Methods}
\label{methods}
\begin{figure*}
    \centering
    \includegraphics[height=4.25cm, width=0.9\linewidth]{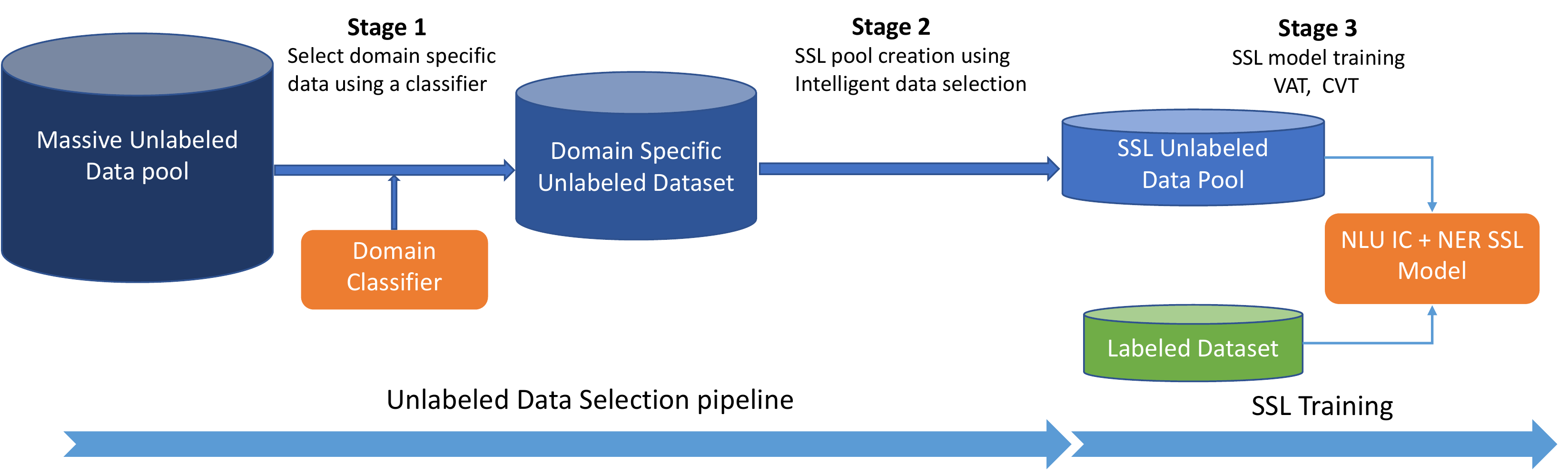}
    \caption{SSL pipeline. Domain specific unlabeled data are first selected using a domain classifer. We then select a subset of the unlabeled data using submodular optimization or committee based selection. Finally we train different SSL models using selected data combined with the labeled data.}
    \label{fig:ssl_pipeline}
\end{figure*}
We are interested in studying two different questions relevant to the use of unlabeled data in production environments: 1) \emph{how to effectively select SSL data from a large pool of unlabeled data}, and 2) \emph{how do SSL techniques perform in realistic scenarios?} 
To do so, we focus on the tasks of intent classification (IC) and named entity recognition (NER), two important components in NLU systems. 


The model architecture we study is an LSTM-based multi-task model for IC and NER tasks, where we use 300-dimension fastText word embeddings~\cite{bojanowski2016enriching}, trained on a large voice assistant corpus.\footnote{The text corpus contains data transcribed by an automatic speech recognition system.} A shared 256-dimension Bi-LSTM encoder and two separate task-specific Bi-LSTM encoders (256-dimension) are applied to encode the sentences. A softmax layer and a conditional random field (CRF) layer are used to produce predictions for IC and NER, respectively.

Below we describe our implementation of the SSL techniques and the data selection methods studied.


%

%

\subsection{Data Selection Approaches}


In the industry setting, we often encounter the situation where we have extremely large pool of unlabeled data, intractable to have SSL methods run on the entire dataset. Given this challenge, we propose a two stage data selection pipeline to create an unlabeled SSL pool, $\mathcal{D}_u$, of a practical size, from the much larger pool of available data.  

Data selection pipeline, shown in Figure \ref{fig:ssl_pipeline}, first uses a classifier's confidence score to filter domain specific unlabeled data from a very large pool of unlabeled data, which might contain data from multiple domains. For a production system, first stage filtering might result in millions of examples, so we further filter data using different selection algorithms to find an SSL data pool, which facilitates effective SSL training. While the first stage filtering tries to find domain specific examples from a large pool, the goal of the second stage filtering is to find a subset of data which could result in better performance in SSL training.  

For first stage filtering, we train a binary classifier on the labelled data, and use it to select the in-domain unlabelled data. In our experiments, switching between different binary classifiers (linear, CNN, LSTM, etc) does not significantly change the selected data. Consequently, in this study, we simply use a single-layer 256-dimension Bi-LSTM for the first stage of filtering. Based on our initial experiments, we use confidence score 0.5 as the threshold for data selection\footnote{We tried  confidence larger than 0.5 but found that a high confidence score degrades the performance. Our hypothesis is that a high confidence score leads to selecting data similar to labeled data hence a less diverse SSL pool.}. For second stage filtering, we explore data selection using a committee of models and using submodular optimization. While this paper explores only two data selection methods, it's worth mentioning that any data selection algorithm can be used in the second stage filtering to further optimize the size of SSL pool.  


\vspace{2mm}

\textbf{Selection by Submodular Optimization:} Submodular data selection is used to select a diverse representative subset of samples from given dataset. This method has been applied in speech recognition~\cite{wei2015submodularity}, machine translation~\cite{kirchhoff2014submodularity} and natural language understanding tasks~\cite{Cho_2019}. For SSL data selection, we use feature-based submodular selection~\cite{kirchhoff2014submodularity}, where submodular functions are given by weighted sums of non-decreasing concave functions applied to modular functions. For SSL data selection, we use 1-4 n-gram as features and logarithm as the concave function. We filter out any n-gram features which appear less than $30$ times in $\mathcal{D}_l \cup \mathcal{D}_u$. The lazy greedy algorithm is used to optimize submodular functions. The algorithm starts with $\mathcal{D}_l$ as the selected data and chooses the utterance from the candidate pool $\mathcal{D}_u$ which provides maximum marginal gain. 

 



\vspace{2mm}

\textbf{Selection by Committee:} SSL techniques work well when the model is able to provide an accurate prediction on unlabeled data. However, when this is not the case, SSL can have a detrimental effect to the overall system, since the model could be creating SSL data that is annotated incorrectly. Ideally, we would like to have a way of detecting when this might be the case.
Typically, for a given input $x$, neural networks provide a point estimate that is interpreted as a probability distribution over labels. If the point $x$ is easy to learn, neural networks trained from different initial conditions will learn a similar probability distribution for $x$. On the other hand, if $x$ is difficult to learn, their predictions are likely to disagree or converge to low confidence predictions. This phenomenon has been observed in several works addressing uncertainty estimation \cite{uncertainty1,uncertainty2}. As a consequence, data points with high uncertainty are more likely to be incorrectly predicted than those with low uncertainty. 

To detect data points on which the model is not reliable, we train a committee of $n$ teacher models (we use $n = 4$ in this paper), and compute the average entropy of the probability distribution for every data point. Specifically, let $P(y;x,\theta_i)$ denote the probability of label $y$ for input $x$ according to the $i^{th}$ teacher, we compute the average entropy of the predicted label distribution of $x$ as: \linebreak $ H(x) = - \frac{1}{n} \sum_{y \in \mathcal{Y}} \sum_{i=1}^n P(y;x,\theta_i) \log P(y;x,\theta_i) $.
We then identify an entropy threshold with an acceptable error rate for mis-annotations (e.g., $20\%$) based on a held-out dataset. Any committee annotated data whose entropy level is higher than the identified threshold, is deemed ``not trustworthy'' and filtered out.




\subsection{Semi-Supervised Learning Approaches}
We explore the following four Semi-Supervised Learning techniques: 

\textbf{PL} based self-training is a simple and straightforward method of SSL~\cite{yarowsky1995unsupervised,mcclosky2006effective}. Using a labeled data set $\mathcal{D}_l$, we first train a ``teacher'' model, $f_\theta$. We then generate a dataset of pseudo-labeled data from $\mathcal{D}_u$, by assigning for each input instance $x_u$, the label $\hat{y}$, predicted by the teacher. A new model, to which we refer as a ``student'', is then trained on the union of both pseudo-labeled and labeled datasets.

In \textbf{KD}, for a given input, a teacher model produces a probability distribution over all possible labels. The predicted probability distribution is often referred to as ``soft label''. The student model is then trained alternating between two objectives: minimizing the loss on the labeled data, defined respectively for different tasks, and minimizing the cross-entropy loss between the student and teacher predicted ``soft label'' on the unlabeled data ~\cite{hinton2015distilling}. The soft labels on intents are generated by the IC's softmax layer, while the soft labels on label sequences are generated per token, by running softmax on the logits for each token before the CRF layer. 

\textbf{VAT} is an efficient SSL approach based on adversarial learning. It has been shown to be highly effective in both image~\cite{vat} and text classification~\cite{vat_classification} tasks. Given an unlabeled instance, VAT generates a small perturbation that would lead to the largest shift on the label distribution predicted by the model. After getting the adversarial perturbation, the objective is to minimize the KL divergence between the label distribution on the original instance and the instance with perturbation. 

\textbf{CVT} is another SSL approach proved to be efficient on text classification, sequence labeling and machine translation~\cite{cvt}. Using an Bi-LSTM, CVT uses the the bi-directional output from current state as an auxiliary prediction, takes the single-directional output from current and neighboring LSTM neurons, and forces them to predict the same label as the auxiliary prediction.

\section{Data Sets}
\label{datasets}

The main motivation of our study is to evaluate different data selection and SSL techniques in a production scale setting where we have a large amount of unlabeled data. To understand impact of data selection, we create two benchmark datasets for our experiments. In both experiments, using the pipeline shown in Figure \ref{fig:ssl_pipeline}, we first select $M$ utterances from a very large pool of unlabeled data, and then apply intelligent data selection to further select $N$ unlabeled utterances.  

\vspace{2mm}

\textbf{Commercial Dataset}: Our commercial dataset provides an experimental setup to compare SSL techniques where \emph{labeled training data and unlabeled data come from a similar distribution}. We choose four representative domains (i.e., categories for which the user can make requests) from a commercially available voice-assistant system for English language. The four selected categories are 1) Communication: queries related to call, messages, 2) Music: queries related to playing music, 3) Notifications: queries related to alarms, timers, and 4) ToDos: queries related to task organization. For each domain, NLU task is to identify the intent (IC), and the entities (NER) in the utterance. \\
For each domain, our dataset contains $50$k unique training, $50$k unique testing utterances, and hundreds of millions of utterances of unlabeled data. Since, we do not know in advance to which domain each unlabeled utterance belongs, we first select $500$K unlabeled utterance per domain to form their respective unlabeled data pool, using a domain classifier, as shown in Figure \ref{fig:ssl_pipeline}. The choice of $500$K size is based on a series of KD based SSL experiments in Music domain, with the SSL data pool size varying from 50K to 1M. It is observed that increasing SSL pool size beyond $500$k starts to reduce the performance gain from SSL (Table \ref{SSL Data Set Size}). To evaluate the effect of intelligent data selection, out of $500$k, we further select $300$k utterances via different data selection approaches and use them as unlabeled data in SSL experiments.

\begin{table}[t]
\caption{\label{SSL Data Set Size} Relative error rate reduction using KD, over baseline trained with only labeled data, for Music domain. Unlabeled data SSL pool size varies from 50K to 1M utterances. 50K labeled examples are used for all experiments. The metric for IC is classification error rate, and for NER is entity recognition F1 error rate. }
\begin{center}
\small{
\begin{tabular}{c|c|c|c|c|c|c}
\hline 
Task & 50K & 100K & 300K & 500K & 1M \\ \hline 
IC & -3.81\% & -3.37\% & -4.40\% & \textbf{-4.49\%} & -4.09\% \\
NER & -6.05\% & -7.49\% & -6.96\% & \textbf{-8.07\%} & -7.20\% \\ \hline

\end{tabular}}
\end{center}
\end{table}

\textbf{SNIPS Dataset}: We also create a benchmark setup where \emph{labeled and unlabeled data come from different distributions}. We use SNIPS~\cite{Coucke2018SnipsVP} dataset as labeled data, and use unlabeled data from our commercial dataset as SSL pool data. Similar to our commercial dataset, we train a binary classifier for each intent on SNIPS and use it to select $300,000$ utterances as the unlabeled data pool for each intent. Then, we apply data selection approaches to filter for $20,000$ utterances per intent for SSL experiments.

\section{Results}
\label{results}

This section presents evaluations of different SSL techniques using different data selection regimes. 
For all experiments, hyperparameters are optimized on development set. The SSL techniques evaluated are: PL, KD, VAT, CVT. The data selection methods evaluated are: random selection (Random), submodular optimization based selection (Submodular), and committee-based selection (Committee). 

\subsection{Results on Commercial Dataset}

\begin{table*}
\small
\caption{\label{SSL Comparison} Error reduction of SSL methods, relative to baseline. \textbf{Bold} represents the best SSL method for a given data selection technique. \textbf{Bold}\textsuperscript{\textdagger} represents the best performance across all SSL methods and data selection techniques.}
\begin{center}
\resizebox{.95\textwidth}{!}{%
\begin{tabular}{c|c|c|c|c|c|c|c|c|c}
\hline 
SSL & Selection & \multicolumn{2}{c|}{Communication} & \multicolumn{2}{c|}{Music} & \multicolumn{2}{c|}{Notifications} & \multicolumn{2}{c}{ToDos} \\ 
\cline{3-10} 
 Algorithm &  Approach & \multicolumn{1}{c|}{IC} & \multicolumn{1}{c|}{NER} & \multicolumn{1}{c|}{IC} & \multicolumn{1}{c|}{NER} & \multicolumn{1}{c|}{IC} & \multicolumn{1}{c|}{NER} &
\multicolumn{1}{c|}{IC} & \multicolumn{1}{c}{NER} \\ \hline 
Baseline &  & 0 & 0 & 0 & 0 & 0 & 0 & 0 & 0 \\ 
PL &  & -3.61\% & -2.86\% & -4.86\% & -3.70\% & -2.79\% & -4.06\% & -2.94\% & -3.33\% \\
KD & \footnotesize{Random} & -6.35\% & -2.97\% & -6.96\% & -4.40\% & -3.48\% & -4.84\% & -4.18\% & -1.59\% \\
VAT &  & -8.14\% & \textbf{-8.18\%} & \textbf{-11.15\%} & \textbf{-9.26\%} & -6.90\% & \textbf{-8.55\%} & -4.07\% & \textbf{-4.59\%} \\
CVT &  & \textbf{-9.61\%} & -5.26\% & -7.21\% & -8.13\% & \textbf{-7.39\%} & -7.19\% & \textbf{-4.75\%} & -2.38\% \\ \hline 

Baseline &  & 0 & 0 & 0 & 0 & 0 & 0 & 0 & 0 \\ 
PL &  & -4.90\% & -3.11\% &-4.61\% & -3.35\% & -1.48\% & -4.62\% & -1.70\% & -4.32\% \\
KD & \footnotesize{Submodular} & -6.69\% & -3.40\% & -8.19\% & -3.63\% & -2.91\% & -4.32\% & -5.01\% & -2.59\% \\
VAT &  & -11.56\% & \textbf{-8.39\%} & \textbf{-14.72\%\textsuperscript{\textdagger}} & \textbf{-11.03\%\textsuperscript{\textdagger}} & -8.70\% & \textbf{-11.86\%\textsuperscript{\textdagger}} & -6.24\% & \textbf{-5.77\%} \\
CVT &  & \textbf{-14.72\%} & -5.91\% & -9.84\% & -9.94\% & \textbf{-8.72\%\textsuperscript{\textdagger}} & -10.61\% & \textbf{-6.30\%} & -3.13\% \\ \hline

Baseline &  & 0 & 0 & 0 & 0 & 0 & 0 & 0 & 0 \\
PL &  & -10.54\% & -3.91\% & -9.02\% & -3.93\% & -6.90\% & -4.47\% & -4.55\% & -3.67\% \\
KD & \footnotesize{Committee} & -11.13\% & -4.46\% & -11.98\% & -4.09\% & -7.76\% & -5.06\% & -6.10\% & -2.61\% \\
VAT &  & -13.16\% & \textbf{-9.40\%\textsuperscript{\textdagger}} & \textbf{-13.63\%} & \textbf{-10.10\%} & -8.50\% & \textbf{-11.82}\% & -5.75\% & \textbf{-5.99\%\textsuperscript{\textdagger}} \\
CVT &  & \textbf{-15.25\%\textsuperscript{\textdagger}} & -6.53\% & -8.72\% & -8.27\% & \textbf{-8.72\%\textsuperscript{\textdagger}} & -10.40\% & \textbf{-7.34\%\textsuperscript{\textdagger}} & -3.58\% \\ \hline 
\end{tabular}}
\end{center}
\end{table*}

Due to confidentiality, 
we could not disclose absolute performance numbers on the commercial dataset. Only relative changes over baseline are reported. A summary of the results for the various data selection and SSL techniques is given in Table \ref{SSL Comparison}. ``Baseline'' refers to model trained with only labeled data. The metric for IC task is intent classification error rate. The metric for NER task is entity recognition F1 error rate. The table shows the relative error reduction compared to baseline. The bold font shows the best performing SSL method for each data selection approach.



\begin{table}[h]
\caption{\label{Full Snips SSL Comparison} Model performance by different SSL methods and data selection methods, for SNIPS data set. The metric for IC task is classification error rate, and for NER task is entity recognition F1 error rate.}
\begin{center}
\small{
\begin{tabular}{c|c|c|c}
\hline 
SSL & Selection & \multicolumn{2}{c}{SNIPS} \\ 
\cline{3-4} 
 Algorithm &  Approach & \multicolumn{1}{c|}{IC} & \multicolumn{1}{c}{NER}  \\ \hline 
Baseline &  & 0.9744 & 0.9367  \\ 
PL &  & 0.9743 & 0.9326  \\
KD & Random & 0.9743 & 0.9424   \\
VAT &  & 0.9814 & \textbf{0.9604}  \\
CVT &  & \textbf{0.9871} & 0.9565 \\ \hline 

Baseline &  & 0.9744 & 0.9367   \\ 
PL &  & 0.9743 & 0.9342  \\
KD & Submodular & \textbf{0.9786} & 0.9403  \\
VAT &  & 0.9728 & \textbf{0.9579}  \\
CVT &  & 0.9785 & 0.9524 \\ \hline 

Baseline &  & 0.9744 & 0.9367   \\ 
PL &  & 0.9700 & 0.9272  \\
KD & Committee & 0.9729 & 0.9353  \\
VAT &  & 0.9772 & 0.9501  \\
CVT &  & \textbf{0.9780} & \textbf{0.9518} \\ \hline 

\end{tabular}}
\end{center}
\end{table}

\textbf{Comparison of Data Selection Methods}: We observe that both Submodular and Committee based selection outperforms random selection across all domains and SSL techniques. This shows the effectiveness of Stage 2 data filtering. While on Notifications and ToDos domain, submodular selection performs better than other methods, on Communication and Music domain, committee based selection performs the best.   

\textbf{Comparison of SSL Techniques}: Table \ref{SSL Comparison} shows that KD improves performances over PL in virtually all scenarios (except for NER in ToDos). This supports the hypothesis that using the full distribution predicted by the teacher model, instead of using solely the predicted label, allows for the transfer of extra information when training a student model. In addition, though both VAT and CVT consistently outperform KD and PL, their benefits are task dependent. VAT shows stronger benefits on all NER experiments, while CVT performs better in most IC experiments. From an accuracy perspective, VAT is more beneficial in NER tasks while CVT is more beneficial in classification tasks. 

\textbf{SSL Techniques Computation Comparison:} We time each SSL technique on the data selected for Music domain. While PL and KD took approximately $30$ minutes to train each epoch on a Tesla V100 GPU, VAT and CVT took $62$ minutes and $75$ minutes, respectively. Given that PL and KD have similar compute requirement and KD consistently outperforms PL, KD should be preferred over PL for SSL. The decision between CVT and VAT relies on the trade-off between accuracy and cost.


\subsection{Results on SNIPS Dataset}

Test results on SNIPS dataset are summarized in Table \ref{Full Snips SSL Comparison}. The test results on SNIPS aligns with our observations on commercial dataset: VAT and CVT are the superior SSL techniques. Moreover, the results show that VAT and CVT provide good generalization even when the labeled and unlabeled data are from different sources and of different distributions. In contrast to the commercial dataset where intelligent data selection leads to better performance, on SNIPS dataset, we found that submodular optimization or committee based selection do not provide any gain over random selection. It's not surprising given that SNIPS labeled data distribution is very different from the unlabeled SSL data which makes data selection algorithm susceptible to noisy unlabeled data selection. For example, submodular optimization primarily optimizes for data diversity which makes it more likely to select diverse unrelated examples than random selection.



\section{Recommendations}
Based on our empirical results, we make the following recommendations for industry scale NLU SSL systems.

\textbf{Prefer VAT and CVT SSL techniques over PL and KL:} When selecting SSL techniques, CVT usually performs better for classification task while VAT is preferable for NER task. In general, we would recommend VAT since its performance in classification task is comparable to CVT and also because VAT excels in NER task which is usually harder to achieve performance gain.

\textbf{Use data selection to select a subset of unlabeled data:} For industry setting where the volume of unlabeled data is impractically large, we introduce a data filtering pipeline to first reduce the size of unlabeled data pool to a manageable size. Our experiments show that both submodular as well as committee based data selection could further improve SSL performance. We recommend Submodular Optimization based data selection in light of its lower cost and similar performance to committee based method. 

From experiments on SNIPS data sets, we observe that further data selection does not bring extra improvement comparing to random selection. Optimizing data selection, when unlabeled data pool is of a drastically different distribution from the labeled data, remains a challenge and could benefit from further research.


\section{Conclusion}
\label{conclusion}
In this paper, we conduct extensive experiments and in-depth analysis of different SSL techniques applied to industry scale NLU tasks. Industrial settings come with some unique challenges such as massive unlabeled data with a mixture of in domain and out of domain data. In order to overcome these challenges, we also investigate different data selection approaches including submodular optimization and committee based filtering.

Our paper provides insights on how to build an efficient and accurate NLU system, utilizing SSL, from different perspectives (e.g. model accuracy, amount of data, training time and cost, etc). By sharing these insights with larger NLP community, we hope that these guideline will be useful for researchers and practitioner who aim to improve NLU systems while minimizing human annotation effort. 

\bibliography{ssl}
\bibliographystyle{acl_natbib}




\end{document}